\documentclass[a4paper,conference]{IEEEtran}
\pdfoutput=1
\usepackage{cite}

\usepackage{booktabs}  

\usepackage[backref]{hyperref}

\ifCLASSINFOpdf
  \usepackage[pdftex]{graphicx}
  \graphicspath{{../pdf/}{../jpeg/}}
  \DeclareGraphicsExtensions{.pdf,.jpeg,.png}
\else
   \usepackage[dvips]{graphicx}
   \graphicspath{{../eps/}}
   \DeclareGraphicsExtensions{.eps}
\fi
\usepackage{amsmath}
\usepackage{algorithmic}

\usepackage{array}

\ifCLASSOPTIONcompsoc
 \usepackage[caption=false,font=normalsize,labelfont=sf,textfont=sf]{subfig}
\else
 \usepackage[caption=false,font=footnotesize]{subfig}
\fi
\usepackage{fixltx2e}
\usepackage{multirow}

\begin{document}
%
\title{Multi-Scale Cascading Network with Compact Feature Learning for RGB-Infrared Person Re-Identification}



%
\author{\IEEEauthorblockN{\IEEEauthorrefmark{1}Can Zhang,
\IEEEauthorrefmark{1}Hong Liu,
\IEEEauthorrefmark{2}Wei Guo ,
\IEEEauthorrefmark{3}Mang Ye}
\IEEEauthorblockA{\IEEEauthorrefmark{1}Key Laboratory of Machine Perception, Peking University\\ Email: \{can.zhang, hongliu\}@pku.edu.cn}
\IEEEauthorblockA{\IEEEauthorrefmark{2}Noah's Ark Lab, Huawei, China\\
Email: guowei67@huawei.com}
\IEEEauthorblockA{\IEEEauthorrefmark{3}Inception Institute of Artificial Intelligence, UAE \\Email: mangye16@gmail.com}}


\maketitle

\begin{abstract}

RGB-Infrared person re-identification (RGB-IR Re-ID) aims to match persons from heterogeneous images captured by visible and thermal cameras, which is of great significance in the surveillance system under poor light conditions. Facing great challenges in complex variances including conventional single-modality and additional inter-modality discrepancies, most of the existing RGB-IR Re-ID methods propose to impose constraints in image level, feature level or a hybrid of both. Despite better performance of hybrid constraints, they are usually implemented with heavy network architecture. As a matter of fact, previous efforts contribute more as pioneering works in new cross-modal Re-ID area while leaving large space for improvement. This can be mainly attributed to: (1) lack of abundant person image pairs from different modalities for training, and (2) scarcity of salient modality-invariant features especially on coarse representations for effective matching. To address these issues, a novel Multi-Scale Part-Aware Cascading framework (MSPAC) is formulated by aggregating multi-scale fine-grained features from part to global in a cascading manner, which results in a unified representation containing rich and enhanced semantic features. Furthermore, a marginal exponential center (MeCen) loss is introduced to jointly eliminate mixed variances from intra- and inter-modal examples. Cross-modality correlations can thus be efficiently explored on salient features for distinctive modality-invariant feature learning. Extensive experiments are conducted to demonstrate that the proposed method outperforms all the state-of-the-art by a large margin.
\end{abstract}

\IEEEpeerreviewmaketitle

\section{Introduction}


Person re-identification (Person Re-ID) 
has witnessed significant progress in computer vision community \cite{52zheng2015scalable}. Early works \cite{68iodice2015salient, 47zheng2012reidentification} mostly rely on designing handcrafted feature descriptors. Recently, deep convolution neural networks (CNNs) are adopted to enhance visual representations, which have dominated benchmarks in person Re-ID.
In addition to inherent challenges such as variances in poses and viewpoints, RGB-based Re-ID relies heavily on good illumination, which greatly degrades its performance in real scenarios due to the complex environments especially during nighttime. 

\begin{figure}[t]
\label{fig_pic}
\centering
\includegraphics[width=2.5in]{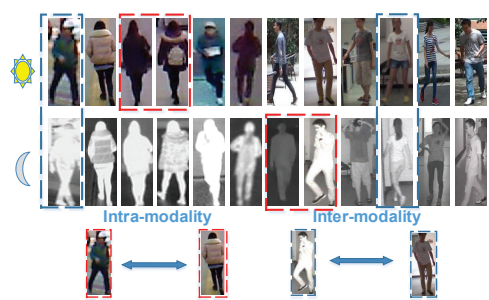}
\caption{Examples of person images selected from public datasets. RGB images $($upper$)$ are captured at daytime by visible cameras while thermal images $($bottom$)$ are captured at night by infrared cameras. Image pairs in each column are of the same identity but different modalities.}
\end{figure}

In order to capture valid appearance information in dark environment, infrared cameras are configured as complementary to visible cameras. In this paper, we mainly focus on a new task proposed recently, namely cross-modality RGB-IR Re-ID, aiming to retrieve query persons in visible (thermal) modality from galleries in another thermal (visible) modality. Some examples of RGB and thermal images are shown in Fig. \ref{fig_pic}, which are captured by visible cameras at daytime and infrared cameras at night respectively. From observations, thermal images enable to provide more invariant person appearances even under poor illumination conditions. As most surveillance cameras can switch between RGB mode and thermal mode automatically \cite{12Wu2017}, RGB-IR Re-ID research has practical significance.

Despite some previous efforts devoted to RGB-IR Re-ID, it still remains challenging and leaves a large space for further explorations. Challenges mainly lie in two folds. Firstly, over-fitting problem results from the limited training data. Making constraints particularly for modality-related variances usually requires abundant training image pairs from different modalities, whereas available datasets are not sufficient enough to support the optimization of network which is expected to narrow the significant gap between modalities. The carefully designed modality-related metrics, such as bi-directional top-ranking (BDTR) loss \cite{15ye2018visible} and its derivative method eBDTR \cite{60ye2020bi-di}, are thus hard to bring about much obvious improvement. Secondly, large modality gap exists due to the inherently homogeneous and heterogeneous variances among RGB and thermal images. Homogeneous variances (\textit{intra-modality variances}) refer to discrepancies within the same modality such as different poses and viewpoints, while \textit{inter-modality variances} is caused by heterogeneous data as RGB and thermal images have different spectrum patterns. It is worth noting that inter-modality variances are always more significant than intra-modality variances, which makes the cross-modality matching task much more challenging. Existing works make constraints for discrepancy elimination from image-level \cite{12Wu2017, 44kang2019person}, feature-level \cite{14Ye2018, 13dai2018cross, 60ye2020bi-di, 61Feng2020learning} and a hybrid of above two \cite{45wang2019learning}. However, these methods fail to explore sufficient and distinctive modality-invariant information simply based on the raw global representations. 

To this end, this paper proposes a  Multi-Scale Part-Aware Cascading framework (MSPAC) to fully exploit sharable information across modalities, and a marginal exponential center loss (MeCen) to explore cross-modality correlations. More specifically, MSPAC is build upon the standard attention mechanism to enhance regions of interest, while in order to leverage discriminative and fine-grained features, we decompose global features into multi-scale parts first, and apply attention module to each single-scale part. To achieve this, attention learning is decomposed into several stages and each stage corresponds to a partition scale for enhancement. In order to extract more deep cross-modality clues, MSPAC unifies the feature representations by aggregating part attentions from small to large scale in a cascading manner. Moreover, based on the unified features, MeCen forces network to learn modality-invariant correlations which enables images of the same identities to cluster compactly in feature space regardless of the modality form. In this case, no additional modality-related constraints are required in RGB-IR Re-ID model, which is conducive to solving the problem of insufficient number of image pairs with different modalities.

The main contributions of this paper are three-fold:

(1) Proposing a multi-scale part aware mechanism to enhance discrimination ability of fine-grained part features in both channel and spatial dimension.

(2) Constructing a hierarchical part aggregation architecture which results in an unified representation by integrating spatial structured information into concatenated local salient features in a cascading fashion.

(3) Introducing a novel MeCen loss to model cross-modality correlations by imposing strong constraints on significant variances which shows superior clustering ability.

\section{Cross-Modality Correlation Via Multi-Scale Part-Aware Cascading Framework}
\noindent\textbf{Problem definition.}
Let $\mathbf{\mathit{V}}=\{v\mid v \in R^{H \times\ W \times 3}\}$ and $\mathbf{\mathit{T}}=\{t\mid t \in R^{H \times\ W \times 1}\}$ denote RGB image set and thermal image set respectively, where $H$ and $W$ are the height and width of the image. Each image $v \in \mathbf{\mathit{V}}$ or $t \in \mathbf{\mathit{T}}$ in dataset has a corresponding label $y \in \{1,2,...,N\}$, here $N$ represents the total number of person identities. A common strategy for deep feature learning is to map images $v$ and $t$ into feature embeddings $x_v=F\{\theta_v ; v\}$ and $x_t=F\{\theta_t ; t\}$, where $F\{\theta\}$ denotes transformation function of the deep network with parameter $\theta$. Given a query image $v$ (in RGB modality) or $t$ (in thermal modality) and a gallery set which is correspondingly constructed by the image $t \in \mathbf{\mathit{T}}$ (in thermal modality) or $v \in \mathbf{\mathit{V}}$ (in RGB modality), the goal of cross modal Re-ID is to retrieve persons with the same identity as query in the gallery set. Feature representations $x_v\in R^d$ and $x_t\in R^d$ are projected into a common feature space by feature embedding module where $d$ is the dimension of feature vectors.

\begin{figure}[t]
\centering
\noindent\makebox[0.5\textwidth][c]{\includegraphics[width=0.48\textwidth]{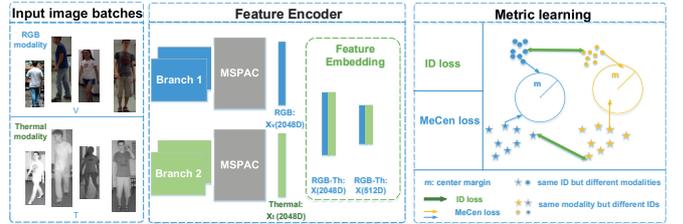}}
\caption{The overall framework of our cross-modality person Re-ID model. There are three components: (1) Input image batches. Each identity in the training batch has the same number of RGB images $v$ and thermal images $t$ randomly selected from datasets. (2) Feature encoding. Two heterogeneous branches (blue and green) and a shared feature embedding module are applied for projecting images into common feature space. (3) Metric learning. Marginal exponential center loss is integrated with identity loss for eliminating complex variances. Geometric figures with different colors and shapes represent differences of identity and modality respectively.}
\label{fig_net1}
\end{figure}

\subsection{Overall Structure}
\label{sec3.1}
The overall structure of the proposed RGB-IR Re-ID model, as illustrated in Fig. \ref{fig_net1}, mainly consists of two modules:

\noindent\textbf{Feature encoder}. The two-branch architecture adopts ResNet50 \cite{21he2016deep} as the backbone, which follows modifications made in \cite{5sun2018beyond}.
Person images $v\in \mathbf{\mathit{V}}$ and $t\in \mathbf{\mathit{V}}$ are 
fed into two heterogeneous branches respectively for deep feature encoding. To eliminate modality-specific noises, multi-scale part aware attention module is designed to incorporate structured semantic information and local salient information into a unified global representations.

\noindent\textbf{Metric learning}. A novel loss, MeCen, is proposed to model cross-modality correlations.
Owing to the superior clustering ability, features of the same identity are pulled closely around the cluster center in the feature space, which leads to an elimination of intra-class variances.

\subsection{Multi-Scale Part-Aware Cascading Framework in Feature Encoder}
\label{MPAC}
The  Multi-Scale Part-Aware Cascading framework (MSPAC) is described in detail in this section, aiming to obtain unified feature representations encoded with both spatial structured information and fine-grained information of the person.

Given the fixed partition scale $s$, as shown in Fig. \ref{fig_net2}, global feature maps from branch $\Theta_v$ (or $\Theta_v$) of the backbone are firstly divided into horizontal stripes. Formally, when the size of global representations is denoted by $H \times W \times D$, the height of part features in scale $s$ is fixed to $\lceil \frac{H}{h_s}\rceil$, where $\lceil \cdot \rceil$ represents the ceiling operation. Suppose there are $k$ partition scales with $s=[s_1, s_2,\cdots, s_k]$, a diversity of granularities can be obtained. To deal with such multiple feature representations, an intuitive motivation is to concatenate all the feature vectors to form a mixed representation or arrange them separately such as the multi-branch architecture in \cite{54wang2018learning}, whereas these strategies either has limited performance or complex network architecture. In comparison, MSPAC jointly integrates local enhanced features into the unified global representations of good discrimination ability in a more simplified manner. The framework is decomposed into two parts for detailed discussions.

\subsubsection*{A. Attention Mechanism with Two-branch Structure}

Attention mechanism is composed of two branches as shown in Fig. \ref{fig_net2}: trunk branch connects original part features from the two-branch backbone network while attention branch contains a spatial attention module and a channel attention module. Spatial attention contributes more to guide the network where to focus and channel attention contributes more to decide what to focus on. Therefore, in order to deal with large intrinsic variances among cross-modal features, both attention structures are adopted to inherit merits of each attention strategy. What's more, as combined attention architecture is computationally efficient, it can be adopted as a plug-and-play module. 

Given feature maps $x_a$ with size of $H \times W \times D$, two different descriptors are generated by average-pooling and max-pooling operation respectively, which are denoted as $x^{ch}_{avg} \in R^{1 \times 1 \times D}$ and $x^{ch}_{max} \in R^{1 \times 1 \times D}$. With only channel dimension, $x^{ch}_{avg}$ and $x^{ch}_{max}$ are further embedded by a shared multiple fully connected network and fused into a unified representations by element-wise summation. In brief, channel attention process is formulated as:
\begin{equation}
\label{eq3} 
CH(x_a)=\sigma(W_F(x^{ch}_{avg}+x^{ch}_{max})+b_F),
\end{equation}
where $\sigma$ denotes the sigmoid function, $W$ and $b$  are the learnable parameters.

The spatial attention module, which concentrates on distinct regions of the body, is introduced as another complementary attention form to the above channel attention module.
Based on two types of pooling operations, feature maps $x^{sp}_{avg}$ and $x^{sp}_{max}$ with size of $H \times W \times 1$ are generated and concatenated along the channel axis. The combined feature maps are encoded into a 2D spatial attention map $SP(x_a^{'}) \in R^{H \times W \times 1}$ by a convolutional layer. Formula of the detailed computation is as follows:
\begin{equation}
\label{eq4} 
SP(x_a^{'})=\sigma\{W_{conv}\{x^{sp}_{avg} \Arrowvert x^{sp}_{max}\}\},
\end{equation}
where $\Arrowvert$ means concatenation operation and $W_{conv}$ denotes parameters in the convolutional layer.

Attention features generated from channel and spatial attention modules are written as the following formula respectively:
\begin{equation}
\label{eq5} 
\left\{
\begin{array}{c}
x_a^{'}=x_a \otimes CH(x_a), \\
x_a^{''}=x_a^{'} \otimes SP(x_a^{'}),
\end{array}
\right.
\end{equation}
where $\otimes$ denotes element-wise multiplication.

In our attention learning architecture, original features are incorporated into part attention maps by residual connection for supervising attention learning. In this case, enhanced feature representations are more robust to unnecessary noises and fatal modifications. Formally, raw features $x_o$ from trunk branch and enhanced features $x_a^{''}$ from attention branch are combined as follows:
\begin{equation}
\label{eq6} 
x=x_o+x_o\otimes x_a^{''}.
\end{equation} 


\subsubsection*{B. Part Feature Aggregation in Cascading Framework}


Considering global features lack fine-grained information and local representations are of better discrimination ability, we partition global features into multi-scale parts for deeply fine-grained information extraction. In this case, clues easily ignored in global representations can be fully exploited and enhanced. Whereas, reliability of part features can be easily affected by the occlusion and pose variances, because the integrity of structured information is inevitably damaged by the rigid partition strategy. To address aforementioned problems, 
a cascading feature aggregation framework is introduced to unify salient information from multiple parts into global representations. In the cascading framework, attention learning endows modality-invariant feature representations with more complete and rich semantic information. Three main components can be summarized, fine-grained feature learning, hierarchical part aggregation and global representation unification. Each component is detailed as follows:

\begin{figure*}[t]
\centering
\noindent\makebox[1\textwidth][c]{\includegraphics[scale=0.15]{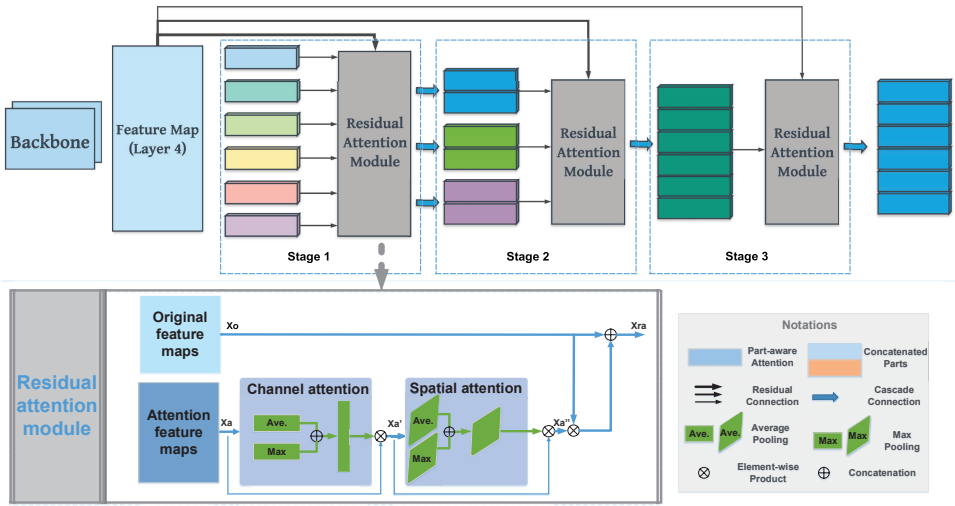}}
\caption{The structure of MSPAC. Upper part: Multi-scale part feature aggregation in cascading framework. Original feature maps are split horizontally into multi-scale parts. In each stage, original features with corresponding partition size are incorporated into attention module by residual connection, denoted as arrow lines. Arrow thickness represents partition size. Bottom part: Attention mechanism with two-branch structure. Attention branch consists of both spatial attention and channel attention where average-pooling and max-pooling operations are included.}
\label{fig_net2}
\end{figure*}



\textit{ Fine-grained feature partition}. Suppose global representations being in the coarsest partition scale of $1$, when the scale is increased, more diverse spatial semantic information can be available from multiple granularities of features. To arrange multiple part features, we denote them as follow:
\begin{equation}
\label{eq1}
X_a= \{x_{a_i, n}^{s_j}\vert i\in [1,r], s_j\in[1,p_i],n\in [1,N]\},
\end{equation}
where $x_{a_i,n}^{s_j}$ is the $s_j$-th part feature in $i$ scale with size of $\lceil\frac{H}{p_i}\rceil \times W \times D$, $N=\sum_{i=1}^rp_i$ denotes that the total number of multi-scale parts, $p_i$ is the number of parts in scale $i$ and $r$ is the total number of partition scales.
 
\textit{ Hierarchical part aggregation}. 
In our cascading framework, diverse granularities of features are incorporated with additional global structured body information and aggregated from part to global to form unified representations. Part features with small scale are combined pairwise with neighbors to form larger representations, and then attention modules in the coarser stage are adopted to further exploit local semantic information. The formulation is written as : 
\begin{equation}
\label{eq2} 
x_{a_m}^{s_j} = F\{\theta_a ;x_{a_{m-1}}^{s_{{i-1}\to j}} \Arrowvert x_{a_{m-1}}^{s_{i\to j}}; x_{o_m}^{s_j}\},  j\in \left[1,2,...,p_r\right],
\end{equation}
where $x_{o_m}^{s_j}$ represents the $s_j$-th original part features introduced by residual connection and $x_{a_m}^{s_j}$ is the corresponding enhanced part features in scale $m$, $\Arrowvert $ denotes the concatenation operation along the height dimension, arrow $\to$ represents the corresponding relation between local features in different stages, \textit{e.g.}, in Eq.\ref{eq2} it indicates that the $j$-th part features with scale $r$ are obtained by the $i$-th part with its neighbor $(i-1)$-th part in previous stage. 



\textit{ Global representation unification}. In the last stage, all the enhanced part features are concatenated together to form the final global representations. As the noises introduced by repeated splitting and combining process can be gradually exaggerated along the optimization chain, in order to alleviate the adverse influence of brutal combination by simply stacking vertically, the initial global features from the backbone are introduced by residual connection. This typical setting of residual learning mechanism provides valid spatial supervision to make separate salient information in concatenated features more complete and structural. The final composite global representation can be formulated as Eq. \ref{eq2} with $p_r=1$.
\subsection{Modeling Cross-Modality Correlation}
As mentioned before, challenges of RGB-IR Re-ID mainly lie in significant cross-modality variances due to the heterogeneous image forms. In order to deal with complex discrepancies across modalities, a novel marginal exponential center loss (MeCen) is presented with expectation of supervising modality-independent feature learning and intra-class variances reduction. 

In cross-modality task, \textit{hard positive examples}, which refer to images with very similar appearances but different identities, mainly comprise images with different modalities. By contrast, images within the same modality compose the easy positive example set due to the smaller discrepancies. Therefore, two key points of the proposed loss lie in, \textit{i.e.}, reducing acceptable variances among easy examples and imposing strong exponential constraints on hard positive examples. The formula can be written as:

 \begin{equation}
\label{eq11} 
\begin{array}{c}
{L_{MeCen}} = {e^{\frac{1}{2}\sum\limits_{i = 1}^n {\max \{ ||{x_i} - {c_{{y_i}}}||_2^2 - m,0\} } }} - 1,\\
s.t.{\kern 1pt} {\kern 1pt} {\kern 1pt} ||c||_2^2 = 1.
\end{array}
\end{equation}
where $||c||_2^2 = 1$ denotes normalization on the center feature vectors.

To eliminate the intra-class accumulation influence, a margin threshold $m$ is introduced. For instance, intra-class variances of feature points, whose distances to clustering centers are less than margin $m$, ought to have no contributions to loss, while most hard positive examples dominate the loss calculation. Furthermore, the clustering ability is further strengthened by imposing restriction power on hard examples. Note that compared with triplet-center loss \cite{62hetriplet-center}, in addition to class-related variances, dealing with modality-related variances among intra-class features is also necessary. As modality-related variances are mostly contained in hard positive examples, the key to improve RGB-IR Re-ID model is closely related to the way in imposing the constraints on such images. 

In our cross-modality Re-ID model, the MeCen loss is introduced to impose strong constraints on intra-class variances while identify loss is mainly used to deal with inter-class variances as complementary. The joint optimization can be defined in a multi-loss function form as follows:
\begin{equation}
\label{eq12} 
L=L_{ID}+\lambda L_{MeCen},
\end{equation}
where $L_{ID}=-\sum_{i=1}^n p_{y_i}log(\hat{p_i})$ and $L_{MeCen}$ denote the identity loss and the MeCen loss respectively, and $\lambda$ is the hyper-parameter to control contributions of the MeCen loss.


\section{Experiments and Discussions}

Extensive experiments are conducted on public datasets to evaluate the performance of the proposed method for cross-modality person Re-ID, which aim to answer the following questions:

\textbf{RQ1:} How does MSPAC-MeCen model perform as compared with traditional methods and state-of-the-art deep learning methods?

\textbf{RQ2:} How do multiple components of the MSPAC-MeCen model contribute to the performance of the overall framework? 

\textbf{RQ3:} How do hyper-parameters introduced by the loss function affect the Re-ID model performance? 

\subsection{Experimental Setting}

\subsubsection*{A. Datasets}
\noindent\textbf{RegDB \cite{33nguyen2017person}} is a small dataset collected by the camera system mounted with both visible and infrared cameras. There are 412 persons in total, each of which has both 10 RGB and 10 thermal images with various gaits. \textbf{SYSU-MM01 \cite{12Wu2017}} contains 491 persons captured by six cameras, including 2 infrared cameras and 4 visible cameras. Training set contains 22258 RGB images and 11909 thermal images of 395 identities, while the remaining 96 identities are divided into query set and gallery set for testing with 3803 thermal images and 301 RGB images respectively. This dataset is collected in both indoor and outdoor environments. All-search mode includes 4 RGB cameras as the gallery set and 2 infrared cameras as the query set, while indoor-search mode excludes two outdoor RGB cameras from the gallery set. Note that our experiments are conducted under the single-shot setting with both two modes for evaluation.

\subsubsection*{B. Evaluation Protocol}
Two standard metrics, cumulative matching characteristics (CMC) and mean average precision (mAP), are adopted to evaluate performance. The detailed CMC rank-k (k=1,10,20) accuracy is listed to make comparison with the existing state-of-the-arts.

For traditional Re-ID models, two commonly used handcrafted feature descriptors, HOG \cite{40danal2005histgram} and LOMO \cite{34liao2015person}, are combined with distance metrics for evaluation, which include KISSEME \cite{36koestinger2012large}, LFDA \cite{37pedagadi2013local}, CCA \cite{38rasiwasia2010new} and CRAFT \cite{30chen2017person}.
The deep learning baselines includes:
TONE \cite{14Ye2018}, Two-stream \cite{12Wu2017}, One-stream \cite{12Wu2017}, Zero-Padding \cite{12Wu2017}, HCML\cite{14Ye2018} which is a two-stage algorithm based on the TONE structure, BDTR \cite{15ye2018visible}, cmGAN \cite{13dai2018cross}, D$^{2}$RL \cite{45wang2019learning}, eBDTR \cite{60ye2020bi-di} which is a derivative method of BDTR model, MSR \cite{61Feng2020learning} and AlignGAN \cite{63wang2019rgb}.

\subsection{Implementation Details}
\noindent\textbf{Experiment settings}. Our network follows parameter settings in \cite{15ye2018visible}. SGD optimizer is applied for optimization with a weight decay of 5e-4 and the momentum term $\beta=0.9$. The initial learning rate is set as 0.01 and decays with 0.1 times every 10 epochs.

\noindent\textbf{Batch sampling strategy}. As the mini-batch sampling strategy \cite{15ye2018visible} 
can not well satisfy clustering prerequisites of MeCen loss. Therefore, this paper revises training strategy with more sample images for each person in a single batch. Specifically, $P$ person identities are randomly chosen in each batch, where each identity has $M$ RGB images and $M$ thermal images. The batch size is represented as $N=2\times P \times M$, e.g., this paper set $M=4$ and $P=8$ for training.

\subsection{Comparison with State-of-the-Art Methods (\textbf{RQ1})}

\renewcommand{\arraystretch}{1.3}
\setlength{\tabcolsep}{2mm}{
	\begin{table*}[t]
		\centering
		\caption{Overall performance comparison with different methods on SYSU-MM01 dataset under all-search and indoor-search mode in terms of rank-k accuracies and mAP (\%). }
		\resizebox{\textwidth}{45mm}{
			\begin{tabular}{c|c|c|c|c|c|c|c|c|c|c|c|c|c|c|c|c|c}
				\toprule
				\multicolumn{2}{c|}{Datasets} & \multicolumn{16}{c}{SYSU} \\
				\hline
				\multirow{3}{*}{Feature} &	\multirow{3}{*}{Metric} &\multicolumn{8}{c|}{All-search}  & \multicolumn{8}{c}{Indoor-search} \\
				\cline{3-18}	
				&&\multicolumn{4}{c|}{Single-shot} &\multicolumn{4}{c|}{Multi-shot} &\multicolumn{4}{c|}{Single-shot} &\multicolumn{4}{c}{Multi-shot}\\
				\cline{3-18}
				&&r=1 & r=10 & r=20 & mAP & r=1 & r=10 & r=20 & mAP& r=1 & r=10 & r=20 & mAP & r=1 & r=10 & r=20 & mAP  \\
				\hline
				\multirow{4}{*}{HOG}
				& KISSME & 2.12 &16.21&29.13&3.53&2.79&18.23&31.25&1.96&3.11&25.47&46.47&7.43&4.10&29.32&50.59&3.61 \\
				& LFDA & 2.33 &18.58&33.38&4.35&3.82&20.48&35.84&2.20&2.44&24.13&45.50&6.87&3.42&25.27&45.11&3.19 \\
				& CCA & 2.74 &18.91&32.51&4.28&3.25&21.82&36.51&2.04&4.38&29.96&50.43&8.70&4.62&34.22&56.28&3.87 \\
				& CRAFT & 2.59 &17.93&31.50&4.24&3.58&22.90&38.59&2.06&3.03&24.07&42.89&7.07&4.16&27.75&47.16&3.17 \\
				\hline
				\multirow{4}{*}{LOMO}
				& KISSME & 2.23 &18.95&32.67&4.05&2.65&20.36&34.78&2.45&3.83&31.09&52.86&8.94&4.46&34.35&58.43&4.93 \\
				& LFDA & 2.89 &21.11&35.36&4.81&3.86&24.01&40.54&2.61&4.81&32.16&52.50&9.56&6.27&36.29&58.11&5.15\\
				& CCA & 2.42 &18.22&32.45&4.19&2.63&19.68&34.82&2.15&4.11&30.60&52.54&8.83&4.86&34.40&57.30&4.47 \\
				& CRAFT & 2.34 &18.70&32.93&4.22&3.03&21.70&37.05&2.13 &3.89&27.55&48.16&8.37&2.45&20.20&38.15&2.69 \\
				\hline
				TONE & HCML & 14.32 &53.16&69.17&16.16&-&-&-&-&20.82&68.86&84.46&26.38 &-&-&-&- \\
				\hline
				\multicolumn{2}{c|}{Two-stream}
				&11.65 &47.99&65.50&12.85&16.33&58.35&74.46&8.03&15.60&61.18&81.02&21.49&22.49&72.22&88.61&13.92 \\
				\multicolumn{2}{c|}{One-stream}
				&12.04 &49.68&66.74&13.67&16.26&58.14&75.05&8.59&16.94&63.55&82.10&22.95&22.62&71.74&87.82&15.04 \\
				\multicolumn{2}{c|}{Zero-Padding}
				&14.80 &54.12&71.33&15.95&19.13&61.40&78.41&10.89&20.58&68.38&85.79&26.92&24.43&75.86&91.32&18.64 \\
				\multicolumn{2}{c|}{BDTR}
				& 27.32 &66.96&81.07&27.32&-&-&-&-&31.92&77.18&89.28&41.86 &-&-&-&-  \\
				\multicolumn{2}{c|}{cmGAN}
				& 26.97 &67.51&80.56&27.80&31.49&72.74&85.01&22.27&31.63&77.23&89.18&42.19&37.00&80.94&92.11&32.76 \\
				\multicolumn{2}{c|}{D$^2$RL}& 28.90 &70.60&82.40&29.20&-&-&-&-&-&-&-&- &-&-&-&- \\
				\multicolumn{2}{c|}{eBDTR}& 27.82 &67.34&81.34&28.42&-&-&-&-&32.46&77.42&89.62&42.46&-&-&-&- \\
				\multicolumn{2}{c|}{MSR}
				& 37.35 &83.40&93.34&38.11&43.86&86.94&95.68&30.48&39.64&89.29&97.66&50.88&46.56&93.57&98.80&40.08 \\
				\multicolumn{2}{c|}{AlignGAN}
				& 42.4 &85.0&93.7&40.7&\textbf{51.5}&\textbf{89.4}&95.7&33.9&45.9&87.6&94.4&54.3&\textbf{57.1}&92.7&97.4&45.3 \\
				\hline
				\multicolumn{2}{c|}{\textbf{MSPAC-MeCen}}& {\textbf{46.62}} &{\textbf{87.59}}&{\textbf{95.77}}&{\textbf{47.26}}&47.57&87.64&\textbf{96.11}&\textbf{38.53}&{\textbf{51.63}}&{\textbf{93.48}}&{\textbf{98.82}}&{\textbf{61.54}}&52.81&\textbf{94.16}&\textbf{99.37}&\textbf{47.09}\\
				\bottomrule
			\end{tabular}
			}
		\label{tab1}
\end{table*}}

This section compares the proposed MSPAC-MeCen with baselines introduced above, including traditional methods and deep learning methods. Table \ref{tab1} and Table \ref{tab2} demonstrate the overall performance evaluated on the SYSU-MM01 and RegDB dataset in terms of Rank$@$K (K=1,10,20) and mAP. From comparison results, the following observations can be derived:


(1) Compared with traditional methods typically based on handcrafted features, among which the highest mAP value even fail to achieve 10\%, MSPAC-MeCen shows superior performance of deep neural network for feature learning.

(2) In all end-to-end deep learning methods, MSPAC-MeCen still achieves attracting performance.
It further verifies that the superiority of the proposed method results from its better ability to obtain distinct features and construct modality-invariant correlations. Moreover, our model need no additional modal-related constraints.

(3) For both all-search and indoor-search mode in Table \ref{tab1}, MSPAC-MeCen model always achieves the best performance with remarkable improvements, which indicates that our model can effectively eliminate background noise influence.

 \setlength{\tabcolsep}{2mm}{
 \begin{table}[t]
     \centering
     \caption{Overall performance comparison with different methods on RegDB dataset in terms of rank-k accuracies and mAP (\%).}
     \begin{tabular}{c|c|c|c|c}
     \toprule
     \multirow{2}{*}{Methods}& \multicolumn{4}{c}{Evaluation Metrics} \\
     \cline{2-5}
      & r=1 & r=10 & r=20 & mAP \\
     \hline
     LOMO & 0.85 &2.47&4.10&2.28 \\
     HOG & 13.49 &33.22&43.66&10.31 \\
     Two-stream & 12.43 &30.36&40.96&13.42 \\
     One-stream & 13.11 &32.98&42.51&14.02 \\
     Zero-Padding & 17.75 &34.21&44.35&18.90 \\
     TONE+HCML & 24.44 &47.53&56.78&20.80 \\
     BDTR & 33.47 &58.42&67.52&31.83 \\
     eBDTR & 34.62 &58.96&68.72&33.46 \\
     MSR & 48.43 &70.32&79.95&48.67 \\
     AlignGAN & 57.9 &-&-&53.6 \\
     \hline
     {\textbf{MSPAC-MeCen}}& {\textbf{49.61}} &{\textbf{72.28}}&{\textbf{80.63}}&{\textbf{53.64}} \\
     \bottomrule
     \end{tabular}
     \label{tab2}
 \end{table}}

\subsection{Ablation Study and Discussions (\textbf{RQ2})}

Individual component of MSPAC-MeCen, including cascading aggregation mechanism and MeCen loss, is further investigated on both datasets
in this section. 
Note our experiments are evaluated in the challenging all-search mode with single-shot setting. 

\subsubsection*{A. Effectiveness of Pyramid Part-aware Attention Mechanism}
\renewcommand{\arraystretch}{1.3}
\setlength{\tabcolsep}{2mm}{
	\begin{table}[t]
		\centering
		\caption{Effectiveness of MSPAC in the proposed cross-modality model on RegDB dataset.}
			\begin{tabular}{c|c|c|c|c|c}
				\toprule
				Methods& r=1 & r=5 & r=10 & r=20 & mAP \\
				\hline
				Baseline+gid & 27.52 &43.54&53.45&63.88&29.44 \\
				MSPAC+gid & 35.58 &49.22&58.74&67.91&36.89 \\
				\hline
				Baseline+pid & 39.08 &52.77&62.48&72.14&42.31 \\
				{\textbf{MSPAC+pid}} & {\textbf{49.61}} &{\textbf{62.77}}&{\textbf{72.28}}&{\textbf{80.63}}&{\textbf{53.64}} \\
				\bottomrule
			\end{tabular}
			\label{tab3}
\end{table}}

\renewcommand{\arraystretch}{1.3}
\setlength{\tabcolsep}{2mm}{
	\begin{table}[t]
		\centering
		\caption{Effectiveness of MSPAC in the proposed cross-modality model on SYSU-MM01 dataset.}
			\begin{tabular}{c|c|c|c|c|c}
				\toprule
				\multirow{2}{*}{Methods}& \multicolumn{4}{c}{Evaluation Metrics} \\
				\cline{2-6}
				& r=1 & r=5& r=10 & r=20 & mAP \\
				\hline
				Baseline+gid & 27.14 &58.77&72.60&84.35&29.22 \\
				MSPAC+gid & 38.86 &68.16&79.78&89.93&40.69 \\
				\hline
				Baseline+pid & 37.97 &73.28&84.33&92.30&41.08 \\
				{\textbf{MSPAC+pid}} & {\textbf{41.41}} &{\textbf{70.87}}&{\textbf{84.09}}&{\textbf{93.66}}&{\textbf{41.98}} \\
				\bottomrule
			\end{tabular}
			\label{tab4}
\end{table}}

\renewcommand{\arraystretch}{1.3}
\setlength{\tabcolsep}{2mm}{
	\begin{table}[t]
 		\centering
		\caption{Comparison of different attention modules. We evaluate the combining methods of spatial attention and channel attention branches, where 3 variants of pooling strategies are further verified, \textit{i.e.}, average pooling, max pooling, and the joint pooling form.}
			\begin{tabular}{c|c|c|c|c}
				\toprule
				\multirow{2}{*}{Methods} & \multicolumn{4}{c}{Evaluation Metrics}\\
		        \cline{2-5}
				& r=1 & r=10 & r=20 & mAP\\
				\hline
				MSPAC \textit{w/o} CH& 42.64 &86.41&93.24&44.57\\ 
				MSPAC \textit{w/o} SP & 40.63 &86.62&95.11&43.97\\ 
				\hline
				Channel \textit{w/o} MP& 42.41 &87.09&96.29&44.56\\
				Channel \textit{w/o} AP  & 42.23 &84.07&94.16&43.40\\
				\hline
				MSPAC-Comb & \textbf{46.62} &\textbf{87.59}&\textbf{95.77}&\textbf{47.26}\\
				\bottomrule
			\end{tabular}
			\label{tab6}
\end{table}}

To validate this,
MSPAC is removed as the baseline for comparison on both datasets in Table \ref{tab3} and Table \ref{tab4}.
As max pooling (MP), average pooling (AP), spatial attention (SP) and channel (CH) attention in MSPAC are all adopted as a joint module, a deep analysis on the advantages of such combination is further made as shown in Table \ref{tab6}. The following conclusions can be drawn:

(1) Compared with the baseline, MSPAC achieves much better performance under both part-based (\textit{pid}) and global-based constraints (\textit{gid}). 
MSPAC with global constraints (\textit{gid}) reach to the similar good performance to the part-based baseline especially on SYSU-MM01 dataset. The unified global representations from MSPAC are thus verified to be of superior discrimination ability.

(2) Feature representations in MSPAC are verified to be discriminative as there is little performance difference between global- and part-level constraints. 

(3) Model usually shows better performance on RegDB dataset due to simple diversities of camera views and background noise.

\subsubsection*{B. Effectiveness of Hierarchical Part Aggregation Architecture}

In order to demonstrate the effect of the hierarchical strategy employed in MSPAC, we compare different configurations and show results in Table \ref{tab7}. Conventional aggregation usually combines multiple single-scale parts into a unified representation where the number of body parts result in different performance, while MSPAC integrates multi-scale body parts in a stage-wise fusion manner. We choose scale value as 3 for experimental verification.

Experiments show our hierarchical fusion yields better performance, which benefits the joint representation from learning multi-scale part-aware features. Besides, a larger part number can lead to better results, \textit{e.g.,} a partition strategy by \textit{P=6} brings more fine-grained information of person appearance which makes it superior to the case \textit{P=3} or \textit{P=1}. Such observation gives us a more reasonable insight in adopting hierarchical aggregation strategy to integrate both coarse and fine level of granularity into a unit.

\renewcommand{\arraystretch}{1.3}
\setlength{\tabcolsep}{2mm}{
	\begin{table}[t]
		\centering
		\caption{Effectiveness of hierarchical part aggregation architecture in the proposed cross-modality model on the SYSU-MM01 dataset.}
			\begin{tabular}{c|c|c|c|c|c}
				\toprule
				\multicolumn{2}{c|}{Methods}& \multicolumn{4}{c}{Evaluation Metrics}\\
				\cline{1-6}
			
				Strategy&Part number& r=1 & r=10 & r=20 & mAP\\
				\hline
				\multirow{3}{*}{Normal}&\{1\} &40.94 &84.22&92.74&44.86\\
				&\{3\} &40.52 &84.28&93.61&42.84\\
				&\{6\}& 42.89&84.91&93.29&44.31 \\
				\hline
				\multirow{3}{*}{Hierarchical}&\{1,3\} & 45.31 &84.70&92.43&44.59 \\
				&\{3,6\} & 33.63 &81.28&92.32&36.65 \\
				&\{1,3,6\} & 46.62 &87.59&95.77&47.26\\
				\bottomrule
			\end{tabular}
			\label{tab7}
\end{table}}

\subsubsection*{C. Effectiveness of Marginal Exponential Center Loss}

\renewcommand{\arraystretch}{1.3}
\setlength{\tabcolsep}{2mm}{
	\begin{table}[t]
		\centering
		\caption{Effectiveness of marginal exponential center loss in the proposed cross-modality model on the SYSU-MM01 dataset.}
			\begin{tabular}{c|c|c|c|c|c}
				\toprule
				\multirow{2}{*}{Methods}& \multicolumn{4}{c}{Evaluation Metrics} \\
				\cline{2-6}
				& r=1 & r=5& r=10 & r=20 & mAP \\
				\hline
				MSPAC+gid & 38.86 &68.16&79.78&89.93&40.69 \\
				\hline
				+center & 40.36 &72.44&83.09&91.22&41.40 \\
				+margin & 42.44 &72.36&82.80&90.98&43.19 \\
				{\textbf{+exp}} & {\textbf{46.62}} &{\textbf{77.20}}&{\textbf{87.59}}&{\textbf{95.77}}&{\textbf{47.26}} \\
				\bottomrule
			\end{tabular}
			\label{tab5}
\end{table}}
The MeCen loss introduces two modifications, margin value for eliminating accumulation influence and exponential form for strengthening the restriction power on hard examples. Table \ref{tab5} shows experiment results conducted on the SYSU-MM-1 dataset to evaluate effectiveness of each modification. 

From observations, incorporating center loss (+center) into the MSPAC model could only have some slight improvements, while further introducing an additional margin $m$ and then even replacing the loss function to an exponential form (+ exp) could improves results with a significant margin in both two terms of evaluation metrics. Therefore, our MeCen loss could achieve significant improvements in eliminating intra-class variances by using the effectiveness of clustering strategy.
\subsubsection*{D. Visualization Results}
The proposed MSPAC-MeCen model is designed to own two main characteristics, robust modality correlations between heterogenous feature representations and superior clustering ability for intra-class discrepancies elimination. Therefore, visualization results are shown in this section to explicitly demonstrate the effectiveness of the proposed method.

\begin{figure}[!t]
\centering
\subfloat[]{\includegraphics[width=0.23\textwidth]{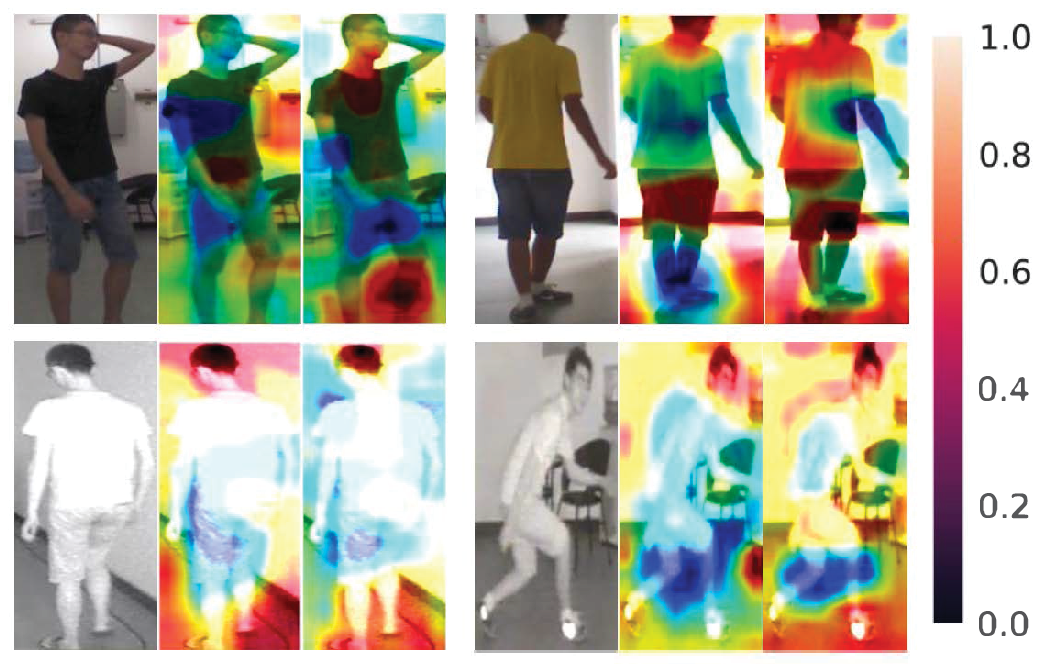}%
\label{fig_att1}}
\hfil
\subfloat[]{\includegraphics[width=0.23\textwidth]{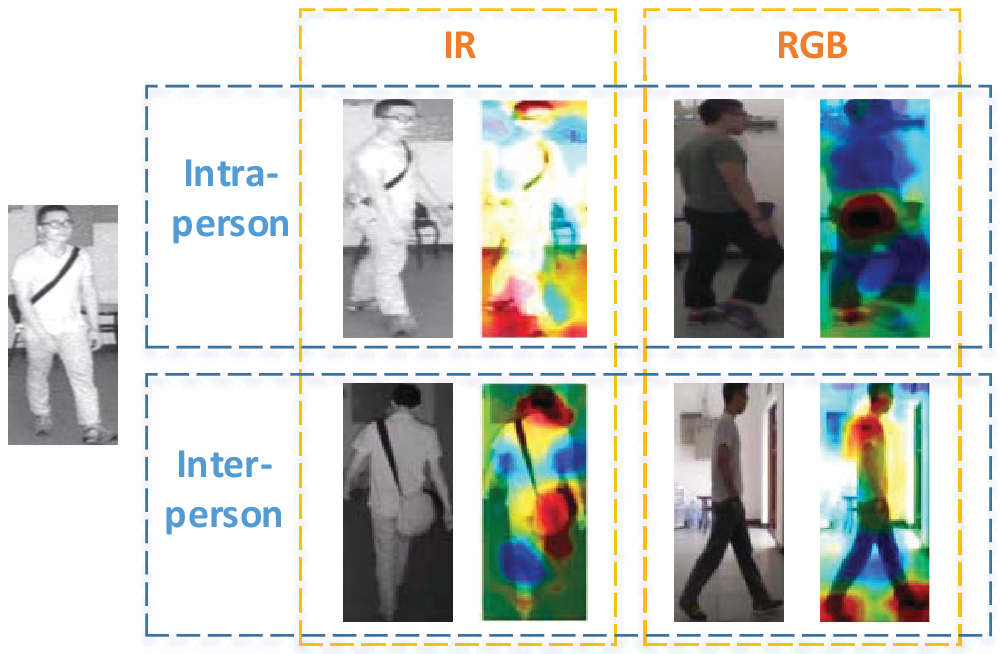}%
\label{fig_att2}}
\caption{Visualization of attention score in heat maps. (a) Examples of persons where each identity corresponds to six images in RGB (upper) and infrared (bottom) modality. Attention maps in middle and right column are results of Baseline+gid model and MSPAC-MeCen model respectively. (b) Positive and hard negative examples of the anchor person. \textit{Intra-person} refers to positive examples of the same identity (upper), and \textit{inter-person} refers to the hard negative examples with similar appearances but different identities (bottom). The attention maps is obtained by the MSPAC-MeCen model.}
\label{fig_attpro}
\end{figure}

Fig.\ref{fig_attpro} shows the attention maps of some example images by means of CAM \cite{55zhou2016learning}, where more salient visual clues will be assigned with higher attention scores. 

To make comparison, RGB images and thermal images of the same person are selected in Fig.\ref{fig_att1}. It can be observed that MSPAC concentrate more on common discriminative regions across modalities such as head and leg regions while baseline method also includes some confusing information like objects in the background. Also more complete semantic information can be obtained by the MSPAC especially under the supervision of the MeCen loss.

\begin{figure}[!t]
\centering
\subfloat[]{\includegraphics[width=0.23\textwidth]{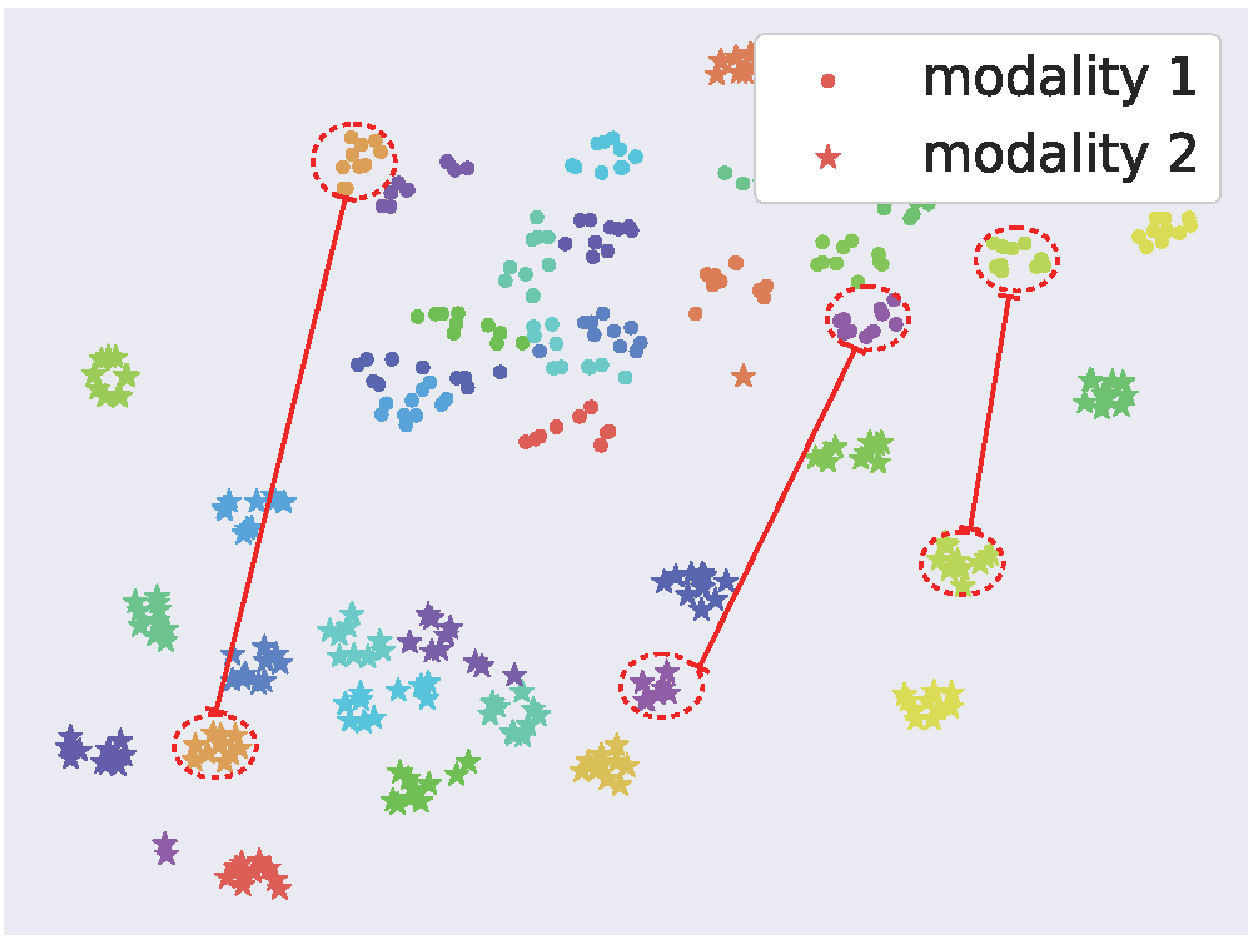}%
\label{fig_clu1}}
\hfil
\subfloat[]{\includegraphics[width=0.23\textwidth]{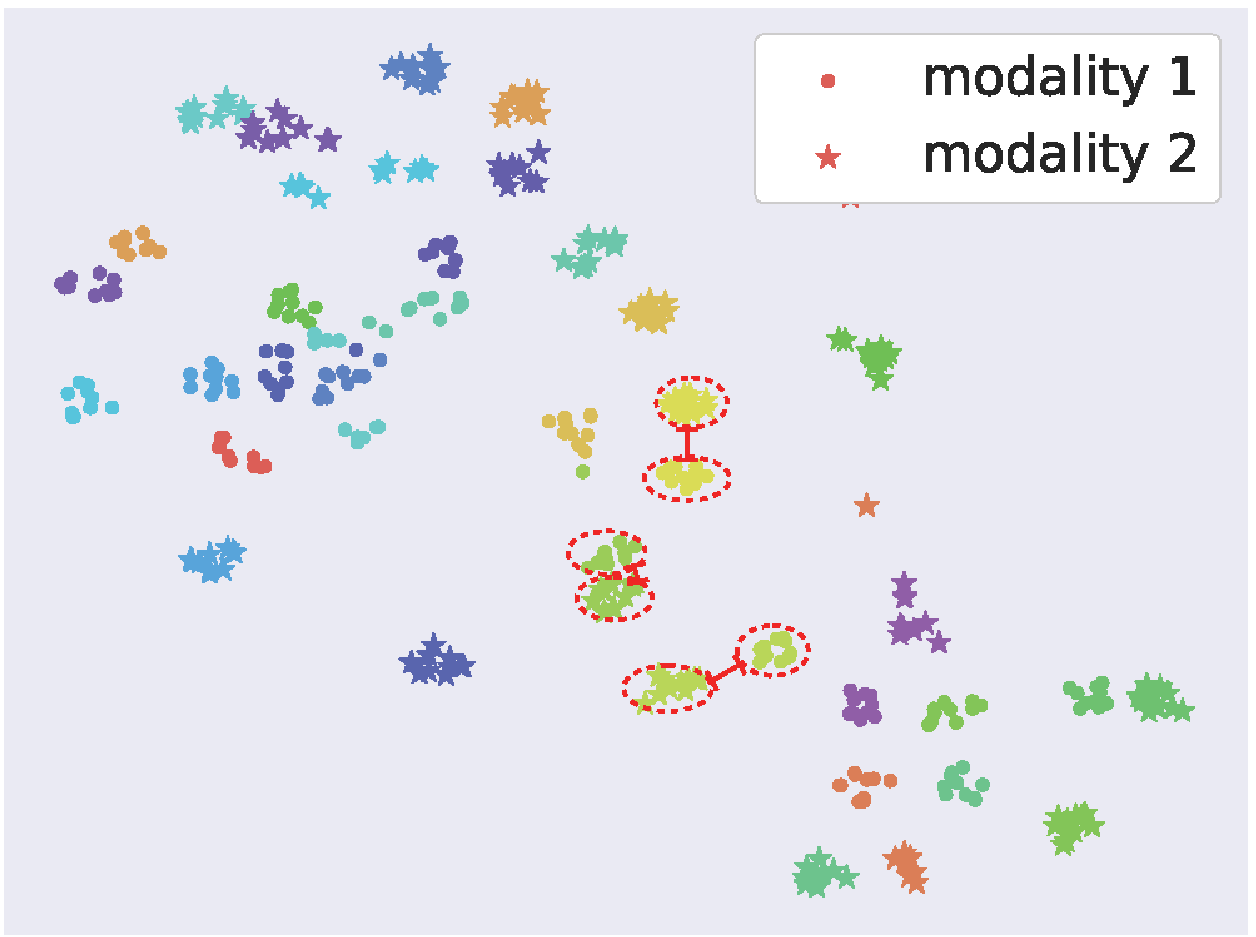}%
\label{fig_clu2}}
\caption{Visualization of clustering performance of MeCen in comparison with the baseline on the SYSU-MM01 dataset. (a) Baseline + gid. (b) The proposed model (MSPAC-MeCen). Only 20 person identities are randomly selected for illustration.}
\label{fig_clu}
\end{figure}

In addition, to explicitly evaluate the clustering ability of the MeCen loss, 20 persons are randomly selected from SYSU-MM01 dataset and their feature vectors are projected into 2D feature space using t-SNE approach \cite{56maaten2008visualizing}. Fig. \ref{fig_clu1} shows distance between modalities in the proposed MSPAC-MeCen model is greatly reduced and feature points are clustered more compact. This demonstrates the superior ability of our model in eliminating modality-related variances.

\subsection{Impact of Hyper-parameter (\textbf{RQ3})}


Two additional hyper-parameters, margin threshold $m$ and weight value $\lambda$, which are introduced in MSPAC-MeCen model, are evaluated in details to give an insight in how to set their values. (Experiment results will be provided in Supplementary Materials due to the paper length limit.)

\section{Conclusions}
For cross-modality Re-ID task, this paper proposes a novel multi-scale part-aware cascading framework (MSPAC) for aggregating sharable features across heterogeneous images and a MeCen loss for modeling cross-modality correlations. The core of our MSPAC is to adopt attention to multi-scale body parts for salient feature enhancement and then incorporate fine-grained features to coarser partition scale step by step in a cascading framework. 
MeCen loss greatly benefits the elimination of intra-class variances and modality-related discrepancies by imposing strong constraints on hard examples for better cross-modality correlations. Extensive experiments on public datasets demonstrate the superiority of the proposed model. Future work involves dynamic weight learning technique for better mutual optimization. Besides, we intend to further investigate effective data augmentations for enlarging dataset.

\section{Acknowledgment}
This work is supported by National Natural Science Foundation of China (No.U1613209), National Natural Science Foundation of Shenzhen (No.JCYJ20190808182209321).
\bibliographystyle{IEEEtran}
\bibliography{MSPAC}
%



\end{document}


%

\section{APPENDIX\footnote{Supplementary material of submission \textbf{Multi-Scale Cascading network with Compact Feature Learning for RGB-Infrared Person Re-Identification}}}
\subsection{Impact of Hyper-parameter (\textbf{RQ3})}
Two additional hyper-parameters, margin threshold $m$ and weight value $\lambda$, are introduced in MSPAC-MeCen model to control the dominant power of accumulation influence over hard examples and MeCen loss over the joint optimization respectively. This section shows how these hyper-parameters influence the Re-ID performance and gives an insight in how to set their value. Note that evaluations are only made on the challenging SYSU-MM01 dataset.

\begin{figure}[!t]
\centering
\includegraphics[width=0.47\textwidth]{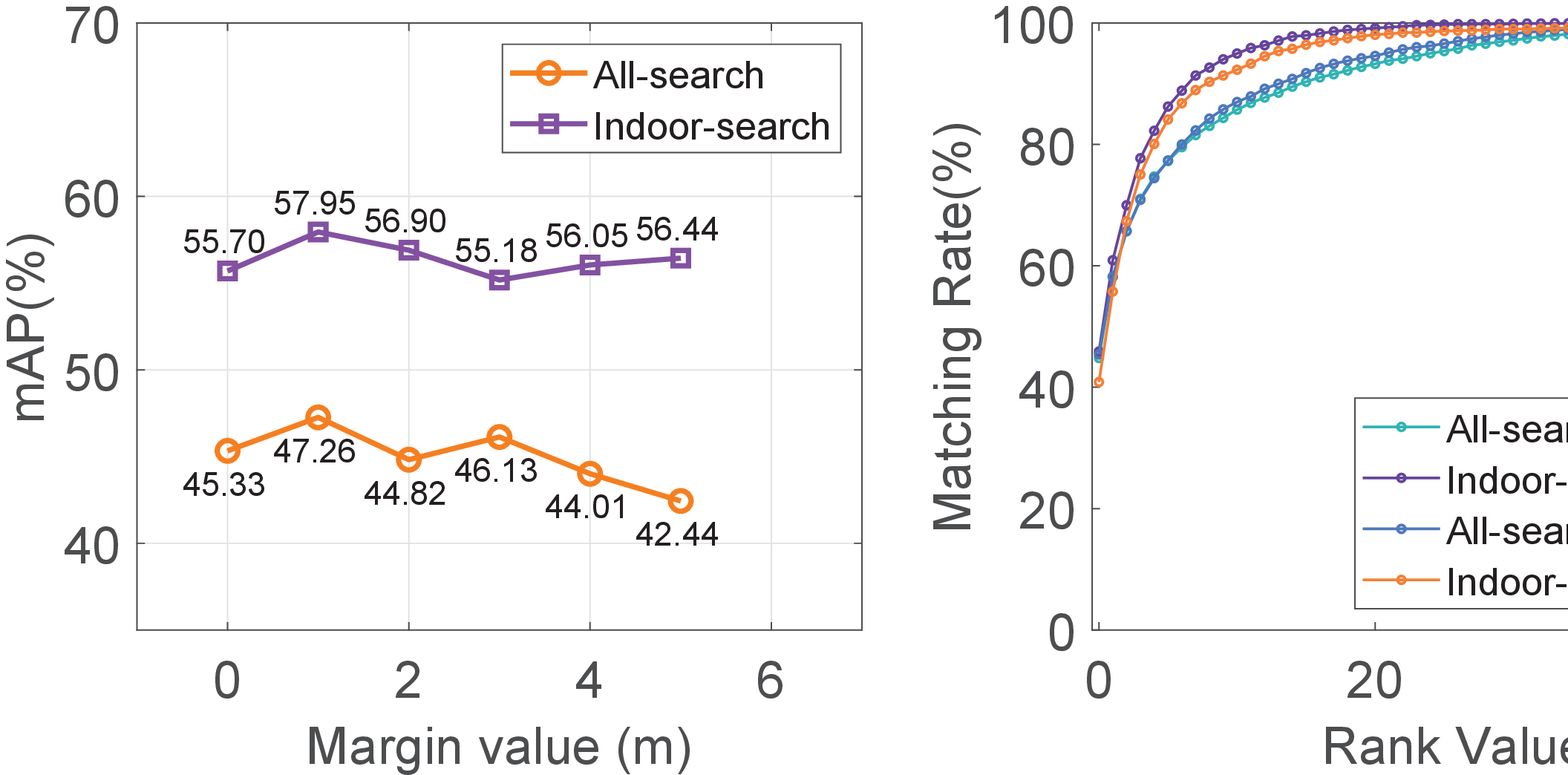}%
\caption{Evaluation of margin value: Models with different $m$ under $\lambda=1$. (Left) mAP metrics in all-search mode and indoor-search mode under different $m$. (Right) CMC curves of the best model with $m=1$ and the bare model with $m=0$.}
\label{fig_mar}
\end{figure}

\begin{figure}[!t]
\centering
\includegraphics[width=0.47\textwidth]{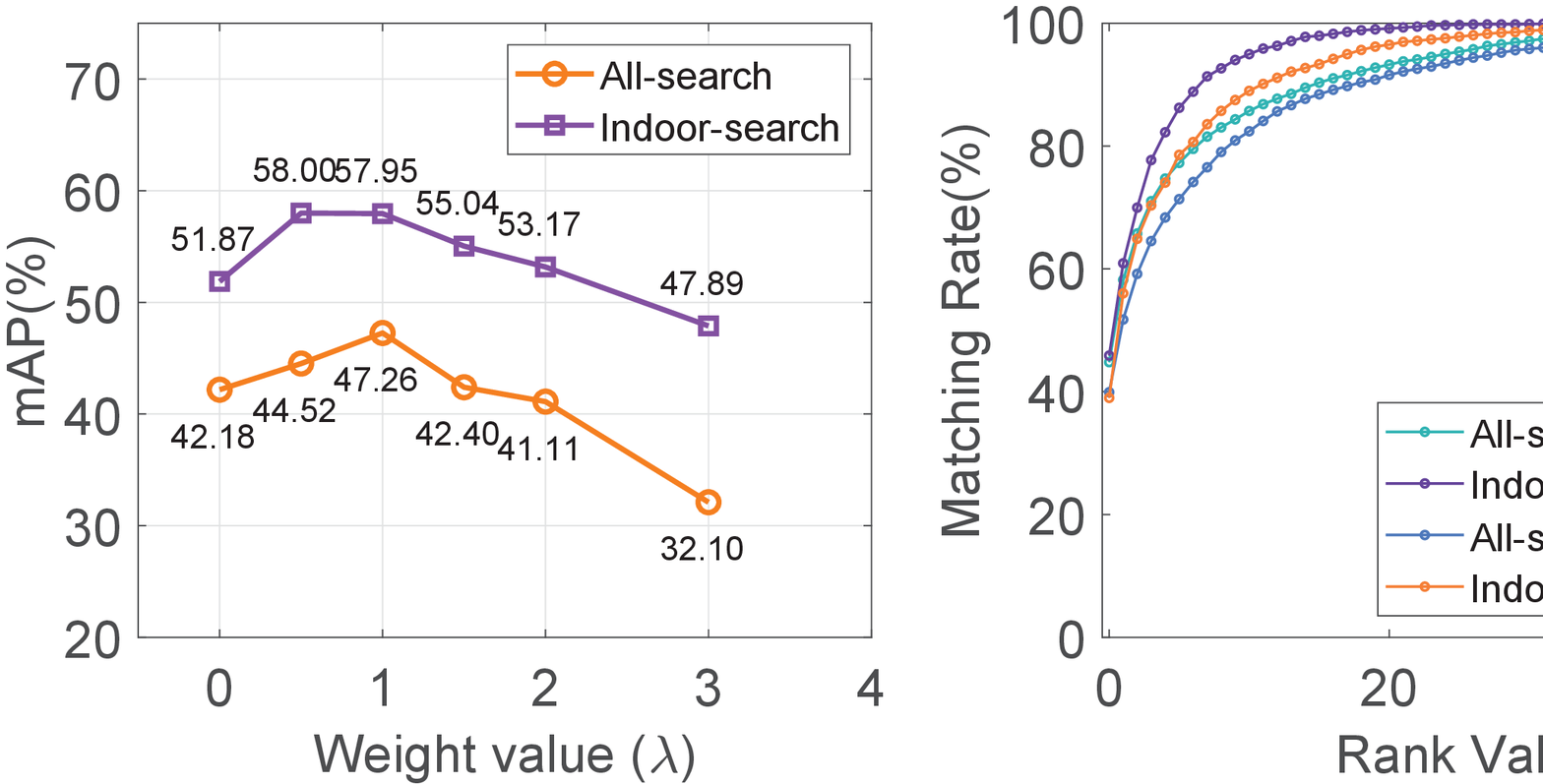}%
\caption{Evaluation of weight value: Models with different $\lambda$ under $m=1$. (Left) mAP metrics in all-search mode and indoor-search mode under different $\lambda$. (Right) CMC curves of the best model with $\lambda=1$ and the bare model with $\lambda=0$.}
\label{fig_wei}
\end{figure}
\subsubsection*{A. Evaluation of margin value}
With $\lambda$ fixed to its optimal value of 1, the margin value $m$ is varied for evaluation. Figure \ref{fig_mar} shows the CMC curves and mAP metric over different $m$ ranging in [0, 1, 2, 3, 4, 5]. MSPAC-MeCen model could achieve the best performance when $m$ is set to 1 in both all-search and indoor-search mode. Compared with $m=0$, there is an improvement of 2.25\% on mAp. It makes sense as margin helps to eliminate the accumulation influence of easy examples. Large cluster size means large intra-class variances will be ignored by the loss function, so increasing $m$ results in a decrease in clustering performance. When $m$ is too large or even exceed the largest feature distance, MeCen loss is reduced to zero and loses clustering ability. Therefore, it is not surprising the poorest performance is demonstrated at around $m\geq 5$ in all-search-mode (Note that it is not a precise point value as our figure is drawn by a series of discrete values with an interval of 1). But what is worth noting is that the changing trend of mAP in indoor-search mode is not significant under different $m$  because indoor-search mode contains much less hard examples by excluding out-door person images with complex backgrounds. These can also be reflected in the CMC curves, where rank matching accuracy in circumstance (m=1) shows slight differences especially in all-search mode compared with no margin case ($m=0$). In all, the introduced margin is verified to be effective in improving Re-ID model performance especially in environment with complex noises. 

\subsubsection*{B. Evaluation of weight value}
Weight value determines how much the MeCen loss dominates in the joint optimization. In order to evaluate its performance, this paper sets $m$ to 1 and varies $\lambda$ in the rang of [0,0.5,1,1.5,2,3] which are much large than those in the original center loss ranging from 0 to 0.1. This difference is due to the normalization of feature vectors assigned by random parameters in initialization. Making trade-off between multiple losses can thus be much easier as sensitivity of weight values to model performance is greatly reduced. In figure \ref{fig_wei}, MSPAC-MeCen model reaches to the highest mAP value at point $\lambda=1$ in the all-search mode. When the weight is set to 0, which means only the ID loss dominates the feature learning, there are some relative drops. It implies that using identity loss simply is not a good choice for learning robust Re-ID model. On the other hand, when $\lambda$ is too large (e.g., larger than 2), Re-ID performance degrades dramatically. \textcolor{blue}{Such observation illustrates that when the identity loss lose the dominant power for network optimization by adopting a larger $\lambda$, centers of clusters are hard to be kept separate from enough distance apart for better discrimination, even though feature points in each single cluster has been pulled compactly around their center. In other words, there exists overlap among different clusters. Therefore, the identity loss is proven to occupy an important position in improving cross-modality person Re-ID performance.} The Although mAP at $\lambda=0.5$ is slightly higher than $\lambda=1$ by 0.05\% in indoor-search mode, as a trade-off for both modes, we set $m$ to 1 in this paper. The CMC curves shown in the right part of the figure demonstrate that matching accuracy changes greatly with different weights, which indicates the importance of the MeCen loss in our MSPAC-MeCen model.